## A Split-face Study of Novel Robotic Prototype vs Human Operator in Skin Rejuvenation Using Q-switched Nd:Yag Laser: Accuracy, Efficacy and Safety

Si Un Chan[a,f*], MD, Cheong Cheong IP[a,g*], MD, Chengxiang Lian[a*], PhD, Muhammad Muddassir[b], MS, Domingo Gomez Dominguez[c], MS, Wai Kit Ming[d], MD, Jianhao Du[d], MPH, Yue Zheng[e#], PhD, David Navarro-Alarcon[b#], PhD, Lie Hua Deng[a#], MD

[a]Department of Dermatology, The First Affiliated Hospital of Jinan University & Jinan University Institute of Dermatology, Guangzhou, China

[b]Department of Mechanical Engineering, The Hong Kong Polytechnic University, Kowloon, Hong Kong

[c]RODS TECHNOLOGY, West Central, Hong Kong Island, Hong Kong

[d]Department of Public Health and Preventive Medicine, School of Medicine, Jinan University, Guangzhou, China

[e]Department of Dermatology and Venereology, The Third Affiliated Hospital of Sun Yat-sen University, Guangzhou, PR China

[f]Department of Dermatology, The University Hospital, Macau University of Science and Technology, Macao

[g]Department of Dermatology, Kiang Wu Hospital, Macao

[*]These authors contributed equally to this article and should be considered as co-first author.

[#]These authors contributed equally to this article and should be considered as formal co-corresponding authors.

**Corresponding author(s):**

1.  Lie Hua Deng, Department of Dermatology, The First Affiliated Hospital of Jinan University & Jinan University Institute of Dermatology, Huangpu Avenue, Guangzhou, Guangdong 510632, China. liehuadeng@126.com
2.  David Navarro-Alarcon, Department of Mechanical Engineering, The Hong Kong Polytechnic University, Hung Hom, Kowloon, Hong Kong. dnavar@polyu.edu.hk
3.  Yue Zheng, Department of Dermatovenereology, The Third Affiliated Hospital of Sun Yat-sen University, 600# Tianhe Road, Guangzhou, Guangdong 510630, China. benbenzhu-11@163.com

**Funding sources**: This research work was supported in part by the Key-Area Research and Development Program of Guangdong Province 2020 (project 76), in part by the Research Grants Council of Hong Kong (grant 14203917), in part by RODS Tech Ltd.

IRB approval status: Reviewed and approved by The Hong Kong Polytechnic University Research Committee; approval #HSEARS20201202001.

Clinicaltrials.gov (or equivalent) listing (if applicable): NCT24435875

**Reprint requests:** Si Un Chan

**Manuscript word count**: 2171
**Abstract word count**: 227
**Capsule summary word count:** 45
**References:** 18
**Figures:** 4
**Tables:** 1

**Abstract ：**

***Background:*** Robotic technologies involved in skin laser are emerging.

***Objective:*** To compare the accuracy, efficacy and safety of novel robotic prototype with human operator in laser operation performance for skin photo-rejuvenation.

***Methods:*** Seventeen subjects were enrolled in a prospective, comparative split-face trial. Q-switch 1064nm laser conducted by the robotic prototype was provided on the right side of the face and that by the professional practitioner on the left. Each subject underwent a single time, one-pass, non-overlapped treatment on an equal size area of the forehead and cheek. Objective assessments included: treatment duration, laser irradiation shots, laser coverage percentage, VISIA parameters, skin temperature and the VAS pain scale.

***Results:*** Average time taken by robotic manipulator was longer than human operator; the average number of irradiation shots of both sides had no significant differences. Laser coverage rate of robotic manipulator ($60.2\pm15.1\%$) was greater than that of human operator ($43.6\pm12.9\%$). The VISIA parameters showed no significant differences between robotic manipulator and human operator. No short or long-term side effects were observed with maximum VAS score of 1 point.

***Limitations***: Only one section of laser treatment was performed.

***Conclusion:*** Laser operation by novel robotic prototype is more reliable, stable and accurate than human operation.

**Capsule summary**

- Robotic prototype effectively, accurately and safely performed the laser operation on facial skin for skin-rejuvenation in this study.

- Robots can replace human beings to perform repetitive laser treatment procedure.

- In the pandemic era, the robotic prototype makes the contactless laser operation possible.

**Introduction**

Laser light energy should be uniformly delivered onto skin tissue, however, there were deficiencies in human operation of lasers [1-3]. Each laser flash lasts only a few nanoseconds and leaves no traces of irradiation on the skin surface, so it is challenging for the human operator to keep track of the irradiated region. The unsteady motion of the operator's hands can also cause missing shots or overdose of laser irradiation [4,5]. Moreover, long-term manual operation leads to higher costs and lower economic benefits. In the pandemic era, patients and operators are at increased risk of exposure to the Covid-19 virus during treatment, which leads to greater demand for contactless medical aesthetics technology.

Robotic technologies involved in skin laser operation and skin surgery are emerging, continuously reforming and innovating. We have developed a novel robotic prototype for skin photo-rejuvenation, which is capable of uniformly delivering laser energy over the skin surface [6] and might help to improve the quality of aesthetic laser treatments.

Q-switched 1064nm Nd:YAG laser has been widely used for facial skin rejuvenation in cosmetic dermatology [7-9].It is popular because of its high efficacy and minimal recovery time, although the clinical outcome varies from one operator to another [7,8,10-12].It is generally believed that non-ablative lasers are safe and its adverse events are usually slight and transient. However, a recent PRISMA compliant systematic review reported that the incident rate of adverse effect in non-ablative lasers was higher than that of exfoliative lasers [13]. One reason for such inconsistent efficacy and high adverse reaction rate may be due to the limitation of human capability in terms of precision, dexterity, repeatability and ability to work in restricted environments [14,15].

In this study, we used the Q-switch 1064nm laser for testing, to compare the accuracy, efficacy and safety of this novel robotic prototype and manual operator in laser operation performance.

## Methods

### Study Subjects

This study included a total of 17 healthy subjects (9 men and 8 women; aged 18-60) with Fitzpatrick skin types II-IV. These subjects were recruited by the Department of Mechanical Engineering, The Hong Kong Polytechnic University, Hong Kong, between February 23, 2019 to May 15, 2021. All subjects were fully informed about the experiment and voluntarily participated in the study. They satisfied all the inclusion and exclusion criteria of the experiment. The candidates were generally healthy and had no physical or mental diseases that could affect the experiment. The exclusion criteria included the following: (1) current skin wound or infection, (2) pregnancy or breastfeeding, (3) any history of keloid or hypertrophic scar, (4) active vitiligo or psoriasis, (5) heavy exposure to the sun, facial energy based treatment, used bleaching products or chemical peeling treatment on the face in the past 4 weeks, (6) hypersensitivity to ultraviolet, (7) history of herpes simplex and/or facial warts, (8) chronic diseases such as lupus erythema, rosacea, solar dermatitis, (9) serious mental disorder, epilepsy, cardiovascular disease, diabetes, immunodeficiency disease, bleeding tendency, or (10) medication regimen that could disturb skin photosensitivity or affect skin regeneration. Candidates were permitted withdrawal from the study for any personal reason. This study was approved by The Hong Kong Polytechnic University Research Committee (approval number: HSEARS20201202001).

### Robotic Laser Prototype

The developed robotic prototype consists of three major components: robot manipulator, robot mobile platform and customized end-effector (Table 1). The facial model was reconstructed through the 3D scanner attached to the robotic arm shown in Figure 1. The reconstructed facial model was recorded in the form of point cloud, where each point contains 3D coordinates, color information and intensity of the reflective surface of the object. Robotic artificial intelligence (AI) would optimize and merge the data from the 3D scanner to reconstruct the accurate facial model. Then the system generates the trajectory of the robotic arm to perform the laser treatment[6].

The human operator could freely select the regions of the face for treatment with the developed graphical user interface and move the mobile robotic platform as per convenience (Figure 2). When corresponding information was input, the robotic end-effector could automatically perform laser treatment at the desired area while bypassing critical facial landmarks such as eyes, lips, eyebrows and hairline.

**Treatment Protocol**

The study was a prospective, comparative clinical trial with a split-face design. Each subject underwent a single-time facial treatment and a follow-up visit after one month. The laser used in this study was the Q-Switched Nd:YAG Laser (K6S, Pusila, CHINA), using a 1,064nm wavelength, 6mm spot size, and energy fluence of 600 mJ/cm$^2$. Two patches of charcoal mask sized 40x50mm$^2$ were applied to both sides of each subject's forehead and two sized of 76×76mm$^2$ on the cheeks. The right side of each participant's face was treated by the robotic prototype and the left side by the human operator. The laser irradiation only performed one pass and each laser irradiation shot should not overlap. The charcoal burned when laser irradiated over the charcoal mask and a trace was left onto the applied surface. Thus the charcoal mask application could serve for a visualization purpose for the observer.

**Outcome Measures**

Digitalized high-resolution photographs (EOS-M3, Canon digital camera, Japan) in combination with the VISIA system (Canfield Imaging Systems, Fairfield, NJ) were used to evaluate outcomes. Images were taken at baseline, immediately after treatment and one month after treatment.

The comparison criteria between the robot and human operators were (1) the time to fill each patch with laser irradiations, (2) the number of laser irradiation shots, (3) the laser covered area (nonoverlapped irradiated area) inside each patch, (4) VISIA data of spots, wrinkles, texture, pores, UV spots, brown spots, (5) the pre- and post-treatment facial temperatures, (6) The VAS (visual analogue scale) painful sensation scores for all of the subjects. (1, 2, 3 was detected by the software of the robotic prototype.)

**Statistical Analysis**

Paired t-test was used to compare all data between both groups. SPSS version 20.0 (IBM Corporation) was used for all statistical analyses. All data are reported as mean ± standard error of the mean. P values of <0.05 were considered statistically significant.

## Results

**Treatment Duration and Laser Irradiation Shots**

The average treatment duration of the robotic manipulator on forehead and cheek was 107.4±23.9 seconds; whereas that of the human operator was 78.6±31.4 seconds. The average time taken by the robotic manipulator was longer than that of human operator both on forehead and cheek（p<0.05）. The average number of irradiation shots of robotic manipulator and the human operator were 267.9±76.7 and 217.1±120.2, which had no statistical differences（p>0.05）.

**Laser Coverage Percentage**

Under the condition of non-overlapping laser emission, the percentage of laser covered area done by robotic manipulator and human operator correspondingly on right and left forehead and cheek patches were compared. The laser coverage percentage of robotic manipulator operation was (60.2±15.1) %, which was greater than that of human operator (43.6±12.9) % (p<0.01). (Figure 3).

**VISIA Comparison**

Evaluation of skin condition was carried out using the VISIA system. The indexes of comparison included: facial spots, wrinkles, texture, pores, UV spots and brown spots. Mean baseline VISIA comparable of the right side and left side showed no significant difference (p >0.05). All indexes were compared one month after treatment against the baseline. The results showed no significant difference between robotic manipulator and human operator (p >0.05). (Figure 4)

**Safety**

All 17 subjects successfully and safely completed the clinical trial without any short or long-term side effects. Subjects' skin temperature was recorded with no statistical differences before and after the laser treatment on the forehead and cheeks by robots and humans（p>0.05）. All subjects tolerated the entire therapeutic procedures well with VAS scores ranging of maximum 1 point.

## Discussion

It is both mentally and physically demanding for human beings to perform repetitive tasks consistently and accurately. With continued future economic and technological development, it will be a waste of resources for humans to perform simple and repetitive laser operations. Robots are designed for these precise and repeatable movements and can perform choreographed tasks smoothly. The physician should become the highly trained operator manipulating the robot, and the two complement each other. Furthermore, in the pandemic era, facing an increasingly large medical aesthetics market, 'contactless medical aesthetics' operated by robots and computers must be the future direction.

The technical challenge of robot-operated laser treatment over the face lies in the complexity of the human facial anatomy. Each individual differs in the curvature of skin surfaces, and the robotic system needs to accurately avoid critical facial landmarks such as the eyes, lips, eyebrows and hairline. This study firstly used robotic system in skin-rejuvenation laser treatment process. The human operator could freely select the specific regions of the face for treatment with the developed graphical user interface and control the mobile robotic platform, and the robotic end effector could automatically perform laser treatment at the desired area while bypassing critical facial landmarks. In this research, the system successfully generated the trajectory of the robotic arm form the reconstructed facial model, so that the distance between the laser emitting head and the skin surface was constantly maintained. This technology is the core to ensure the safety of the robotic manipulator in facial rejuvenation treatment. In all of our subjects, there was not a single case of false targeting of coarse hair or vital organs; and no side effects such as blisters, hyperpigmentation or hypopigmentation were reported. The degree of pain in all subjects and the temperature of the facial skin before and after the laser treatment can also reflect its safety to a certain extent.

In Lim HW's study, the efficacy between novel robot-assisted laser hair removal and physician-directed hair removal were also compared [16]. We found that the novel robotic system was superior to human operator in terms of laser coverage rate and accuracy of non-repetitive emission, and similar conclusions reached. This advantage of the novel robotic system helps avoid laser omission and overdose which often occurs in human-operated laser treatment. Improvement of facial skin rejuvenation could be evenly presented, and side effects caused by excessive local energy such as blisters, pigmentation, or discoloration could be minimized.

The efficacy of Q-switch 1064nm laser in skin rejuvenation has long been proven [10-12]. In the actual treatment of skin rejuvenation, Q-switch 1064nm laser generally requires 3-5 passes until a clinical end point of pinpoint bleeding is observed to achieve a satisfactory treatment effect. In this study, we compared the skin spots, wrinkles, skin texture, skin pores, UV spots and brown spots of the subjects before and one month after the laser treatment using the VISIA Complexion Analysis, and we found that there was no significant difference in the indicators of the two, showing that the effects of a one-pass laser treatment by the robotic manipulator and the human operator were similar. Focusing on the laser coverage rate under one-pass laser irradiation instead of multi-pass treatment, we found that the efficacy of treatment end point was not observed in VISIA. Considering the results of previous published research [11,17,18], it is reasonable to speculate that whether it is one-pass or multi-pass treatment, the laser emitted by the robotic prototype will act on the skin more uniformly, thus the resulting curative effect would be similar to or better than that of manual operation. In addition, whether the application of topical carbon solution before laser treatment might improve the efficacy of facial rejuvenation is still controversial [9].

The average operation time taken by the robotic platform was longer than that of the human operator in this study. One of the reasons was that the number of laser irradiation shots carried out by the robotic platform were greater; conversely, when we tried to increase the speed of the robotic arm, the shots may be fired less precisely. In Lim's research, within a skin area of 12cm*9cm, single-pass laser irradiation, the average treatment duration and number of irradiation shots administered by the automatic robot system were 18 minutes, 30 seconds for 260 shots [16]. In comparison, the operating efficiency of our robotic platform had been significantly improved. Therefore, it would be a better choice to extend the operation time appropriately to ensure the accuracy of the robotic performance.

Our result confirmed that this robotic prototype can establish a stable launch angle, frequency and coverage rate during skin photo-rejuvenation procedures, thus achieving safer and more accurate results than dermatologists. This research is expected to provide a foundation for future research and development in this field. Since the robotic platform can be connected to a variety of laser instruments, it could be widely used in different skin laser treatments in the future. Further optimization of the treatment speed would be addressed following the introduction of commercial research and development.

## Conclusion

The operation of laser by robotic prototype is more reliable, stable and accurate than human operator. Our research provides novel insights for skin resurfacing technology and might help to further understand the robotic laser prototype in dermatology. It might offer a foundation for the future research in artificial intelligence for diagnosis and treatment of skin problems.

**Abbreviations used:**

AI, artificial intelligence

**Figure legend**

Figure 1.   Computer Generated Image of 3D point cloud data.  The user interface showed the selected regions of the face for treatment with thermal visualization.

Figure 2.  The proposed robotic prototype.

Figure 3.  The comparison of laser coverage rate between robotic prototype and human practitioner in two subjects. U represented the total operable area, Φ the covered area and (U-Φ) the uncovered area.

Figure 4. Quantitative analysis of facial spots, wrinkles, texture, pores, UV spots and brown spots by VISIA system. The results showed no significant difference between the robotic prototype and human operator.

**Table legend**

TABLE 1. SUMMARY OF THE ROBOTIC LASER SYSTEM SPECIFICATIONS

| Device | Model, Manufacturer | Size (weidth x length x height) | Specification |
|---|---|---|---|
| Dermatological Laser | Q-switch 1064nm Nd:YAG, (K6S, Pusila, CHINA) | 40×32×34cm$^3$ | Voltage: 220V. Wavelengh:532/1064/1320nm. Control: Q-switch |
| Robot Arm | UR5$^{TM}$, Universal Robots ® | Maximum (extend): 850 mm | Payload:5kg Weight: 18.4kg |
| Structured-light 3D Camera | Astra Embedded S$^{TM}$, Orbbec ® | 68.6x22.3x14.8 mm$^3$ | Range:0.25-1.5m Depth FOV: H67.9°,V45.3°, D78° ±3.0° RGB FOV: H71.5°,V56.7°, D84° |
| Thermal Sensor | Board: Purethermal 2$^{TM}$, GroupGets® Lens:Lepton® 3.5, FLIR® | 30x22x13 mm$^3$ | Frequency: 9Hz Resolution:160 x120 pixels Range Temperature Models: -10 to 140° C and -10 to 450° C |
| Computer | Self-Assembled | 30x30x20 mm$^3$(approx..) | Intel i7, 12GBRAM. |

Fig. 1 (a)Computer Generated Image of 3D point cloud data. (b)The user interface showed the selected regions of the face for treatment with thermal visualization. The green circle showed how the skin temperature was measured by the thermal sensor.

(a)

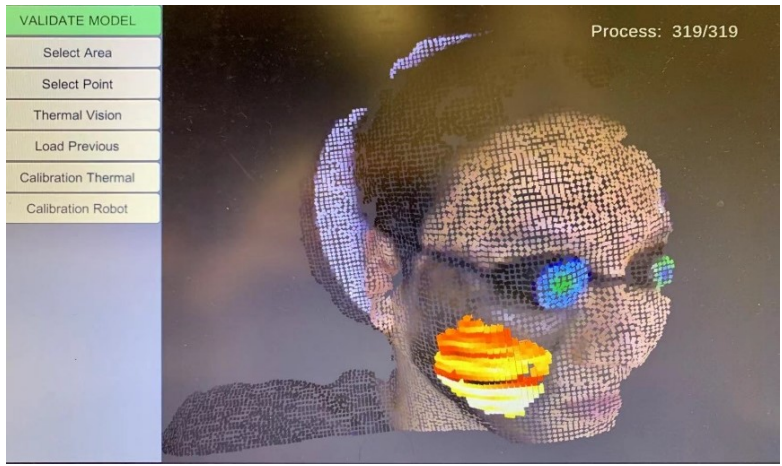

(b)

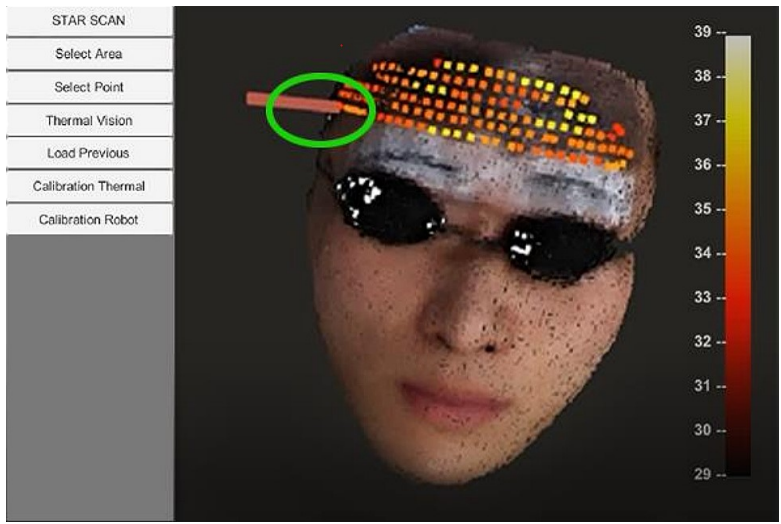

Fig. 2 (a) The proposed robotic prototype.

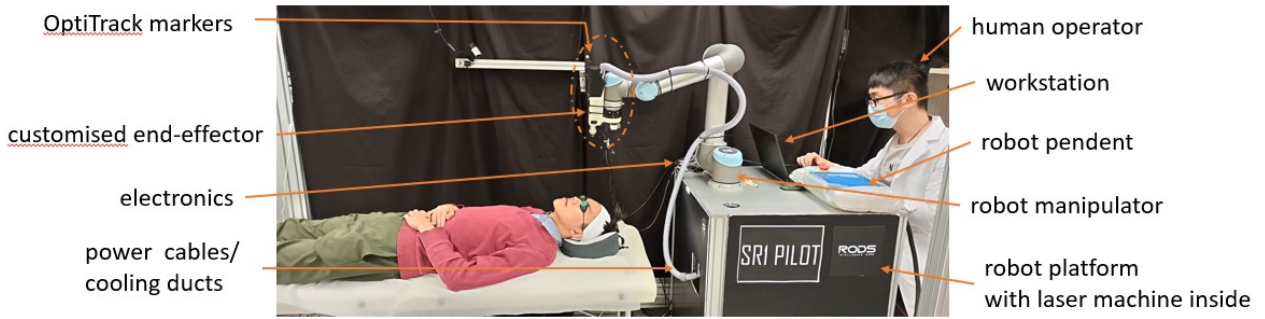

(b) The end effector of the robotic prototype.

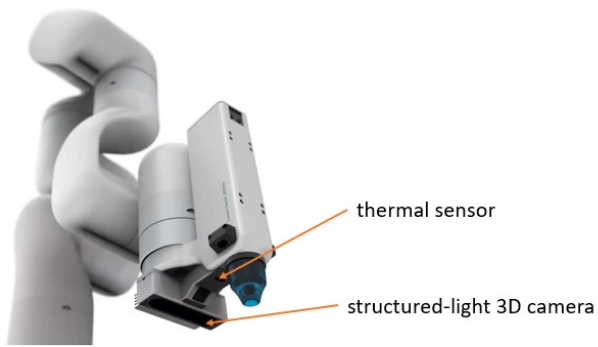

Fig. 3: The comparison of laser coverage rate between robotic prototype and human practitioner in two subjects. U represented the total operable area, Φ the covered area and (U-Φ) the uncovered area.

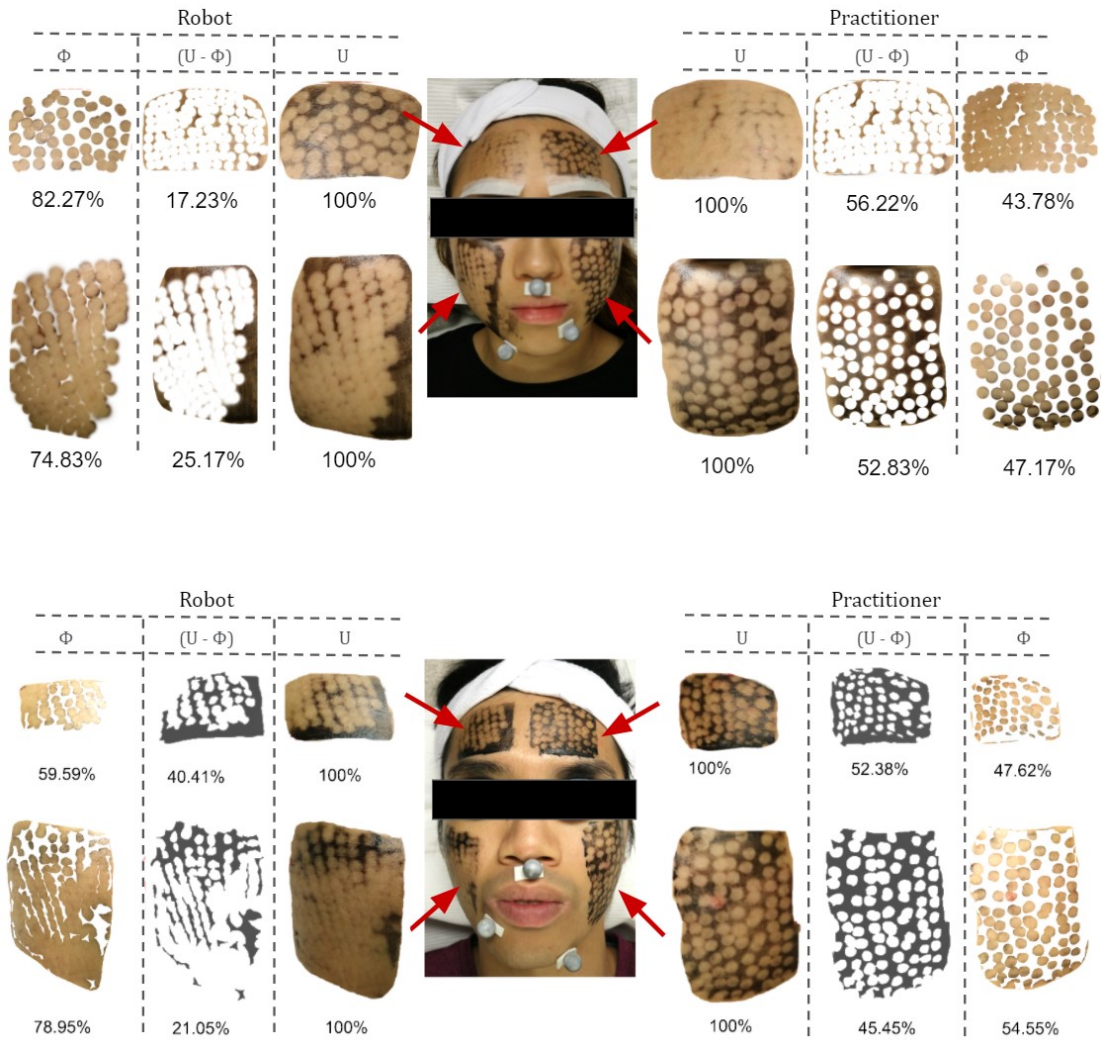

Fig. 4：Quantitative analysis of facial spots, wrinkles, texture, pores, UV spots and brown spots by VISIA system. The results showed no significant difference between the robotic prototype and human operator.

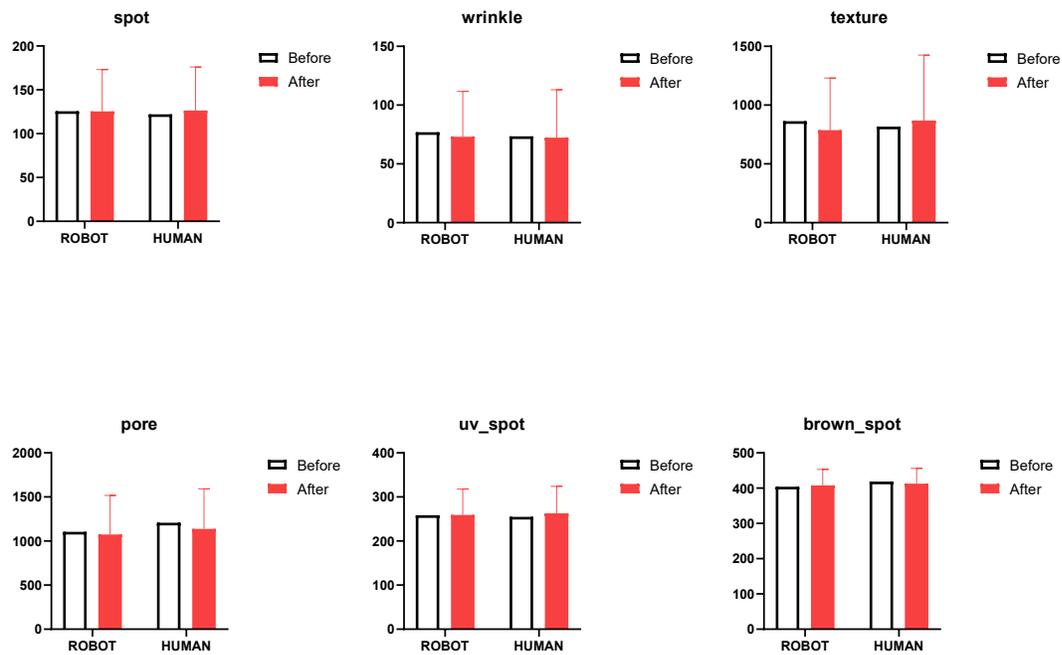